\documentclass[runningheads]{llncs}
\usepackage[T1]{fontenc}
\usepackage{graphicx}
\usepackage{subcaption}
\usepackage{multirow}
\usepackage{float}
\usepackage{censor}

\begin{document}
\title{Revisiting N-CNN for Clinical Practice}
%
%
\author{Leonardo Antunes Ferreira\inst{1}\orcidID{0000-0002-9856-2460} \and
Lucas Pereira Carlini\inst{1}\orcidID{0000-0003-0999-186X} \and
Gabriel de Almeida Sá Coutrin\inst{1}\orcidID{0000-0002-9211-6101} \and
Tatiany Marcondes Heideirich\inst{1}\orcidID{0000-0003-2038-7719} \and
Marina Carvalho de Moraes Barros\inst{2}\orcidID{0000-0001-6989-3474} \and
Ruth Guinsburg\inst{2}\orcidID{0000-0003-1967-9861} \and
Carlos Eduardo Thomaz\inst{1}\orcidID{0000-0001-5566-1963}}

\authorrunning{L.A. Ferreira et al.}

%
\institute{FEI University Center, Sao Bernardo do Campo SP 09850-901, Brazil \and
Federal University of Sao Paulo, Sao Paulo SP 04024-002, Brazil \\
\email{\{leferr,cet\}@fei.edu.br}}
\maketitle              
\begin{abstract}
This paper revisits the Neonatal Convolutional Neural Network (N-CNN) by optimizing its hyperparameters and evaluating how they affect its classification metrics, explainability and reliability, discussing their potential impact in clinical practice. We have chosen hyperparameters that do not modify the original N-CNN architecture, but mainly modify its learning rate and training regularization. The optimization was done by evaluating the improvement in F1 Score for each hyperparameter individually, and the best hyperparameters were chosen to create a Tuned N-CNN. We also applied soft labels derived from the Neonatal Facial Coding System, proposing a novel approach for training facial expression classification models for neonatal pain assessment. Interestingly, while the Tuned N-CNN results point towards improvements in classification metrics and explainability, these improvements did not directly translate to calibration performance. We believe that such insights might have the potential to contribute to the development of more reliable pain evaluation tools for newborns, aiding healthcare professionals in delivering appropriate interventions and improving patient outcomes.


\end{abstract}


\section{Introduction}
In recent years, Artificial Intelligence (AI) models capable of automatically detecting pain through facial expression analysis have gained significant attention, addressing the issues of human subjectivity and untreated pain \cite{gkikas2023automatic}. Among these models, the Neonatal Convolutional Neural Network (N-CNN) stands out as the first end-to-end Deep Learning model specifically designed and trained for neonatal pain detection based on facial expressions \cite{zamzmi2019pain}. In fact, the N-CNN  offers a lightweight architecture, that requires less memory and lower computational resources, making it suitable for embedding into mobile or clinical devices within neonatal intensive care units (NICU), ensuring faster training and recall capabilities.

Models intended for clinical practice must exhibit qualities such as accuracy, explainability and reliability \cite{gkikas2023automatic,jiang2012calibrating,kompa2021second}. While the N-CNN has already proven to be accurate \cite{zamzmi2019pain}, limited research has been conducted regarding its hyperparameters, explainability and reliability. Hyperparameter optimization, or tuning, is often used to improve model performance without altering its architecture. Although this performance improvement can be verified by traditional classification metrics and eXplainable Artificial Intelligence (XAI) methods, less attention has been given to model reliability. Being reliable, or calibrated, is a highly desirable property in automatic medical diagnosis, as it enables the estimation of associated risks during clinical practice \cite{jiang2012calibrating,kompa2021second}. For instance, high-confidence predictions made by an AI model should exactly match the likelihood of that event occurring, otherwise, overconfident predictions can potentially lead to harmful consequences.  In this context, this paper investigates the effects of hyperparameter optimization on classification metrics,
explainability and reliability discussing their potential impact in clinical practice. 
\section{Materials and Methods}
For training and testing the N-CNN, we utilized two facial image datasets: iCOPE \cite{brahnam2006machine} and UNIFESP \cite{heiderich2015neonatal}. In this way, the model was trained from scratch. The iCOPE dataset comprised 60 images labeled as \textit{{“Pain”}} and 63 images labeled as \textit{“No Pain”} from 26 newborns (NBs). The UNIFESP dataset included 164 \textit{“Pain”} images and 196 \textit{“No Pain”} images from 30 NBs. To ensure robust evaluation, we employed a leave-sample-subjects-out cross-validation method \cite{coutrin2022} with 10 folds, using the same 10 folds in all evaluations. Data augmentation was applied to training images in each fold, generating 20 new images with varied parameters such as width and height shift (0.20), rotation (30º), shear (0.15), brightness (0.50 – 1.10), zoom (0.70 – 1.50), and horizontal flip.

\subsection{N-CNN Performance Evaluation}

We selected hyperparameters according to Table \ref{hyper_table} that do not modify the original N-CNN architecture \cite{zamzmi2019pain}, primarily affecting learning rate and training regularization. However, a slight modification was required in the N-CNN output activation function to accommodate the Label Smoothing Regularizer (LSR) \cite{szegedy2016rethinking} applied here. Specifically, we replaced the single output neuron activated by a sigmoid function with two neurons activated by a softmax function, being each neuron responsible for providing normalized probabilities for the respective classes. We exclusively focused on the \textit{“Pain”} neuron output, where a threshold of 50\% was set to consider a prediction belonging to this class.

Since we separately evaluated each hyperparameter, only those that demonstrated an improvement in F1 Score were selected to make the Tuned N-CNN. To assess the performance of the final model, we employed standard evaluation metrics including Accuracy, F1 Score, Precision, and Sensibility. We also used Grad-CAM ($GC$) \cite{selvaraju2017grad} and Integrated Gradients ($IG$) \cite{sundararajan2017axiomatic}. Both are considered to be attribution methods and provide further insights into the model's decision-making process, but $GC$ focuses on regions of the images deemed discriminant to classification, whereas $IG$ assigns to each pixel an importance value to the final classification. Regarding the $GC$, we generated the attribution mask based on the last convolutional layer of the N-CNN, where the model should capture high-level features and semantic information \cite{selvaraju2017grad}.

Additionally, we employed calibration curves to validate whether the confidence scores yielded by the N-CNN are aligned with the frequency of the \textit{“Pain”} class occurrences. The calibration curve plots the frequency of the positive class against the model's confidence scores. A perfectly calibrated model is represented by a 45º diagonal line, that is, the confidence is exactly equal to the actual frequency of events \cite{guo2017calibration}. Points below the diagonal indicate overconfidence, where the model assigns higher probabilities to the occurrence of events than what is observed in reality. Conversely, points above the diagonal indicate a lack of confidence, where the model's confidence is lower than expected despite correctly identifying most of the positive class samples. To quantify if a model is calibrated, we used the Expected Calibration Error (ECE) \cite{naeini2015obtaining}. This metric divides predictions confidences into $K$ equally-spaced bins, here $K=10$, and calculates the weighted average of the differences between the frequency of the positive class and the confidence within each bin, providing a quantification of the model's miscalibration \cite{naeini2015obtaining}. It is important to note that the calibration curve focuses solely in the positive class, considered here as the \textit{“Pain”} class.

\subsection{NFCS as Soft Label}

One of the studied hyperparameters is the value $\epsilon$ of the LSR \cite{szegedy2016rethinking}. It involves adjusting the ground truth labels during training by introducing a small amount of smoothing. Instead of assigning a hard label (binary target) of 0 or 1 to a particular class, LSR assigns a slightly softened label (non-binary target), such as [0.1, 0.9]. This regularization technique helps prevent the model from becoming overconfident and overfitting \cite{szegedy2016rethinking}. Nonetheless, applying the same smoothing value for all labels can potentially result in information loss as we increase $\epsilon$, encouraging the model to be less confident about the true class. Therefore, we propose a novel approach for establishing independent class membership probabilities for each image using the Neonatal Facial Coding System (NFCS) \cite{grunau1987pain}, as it is widely used and available in the UNIFESP dataset \cite{heiderich2015neonatal}.  

\begin{figure}[!t]
  \centering
  \begin{subfigure}[b]{0.47\textwidth}
    \centering
    \includegraphics[width=\textwidth]{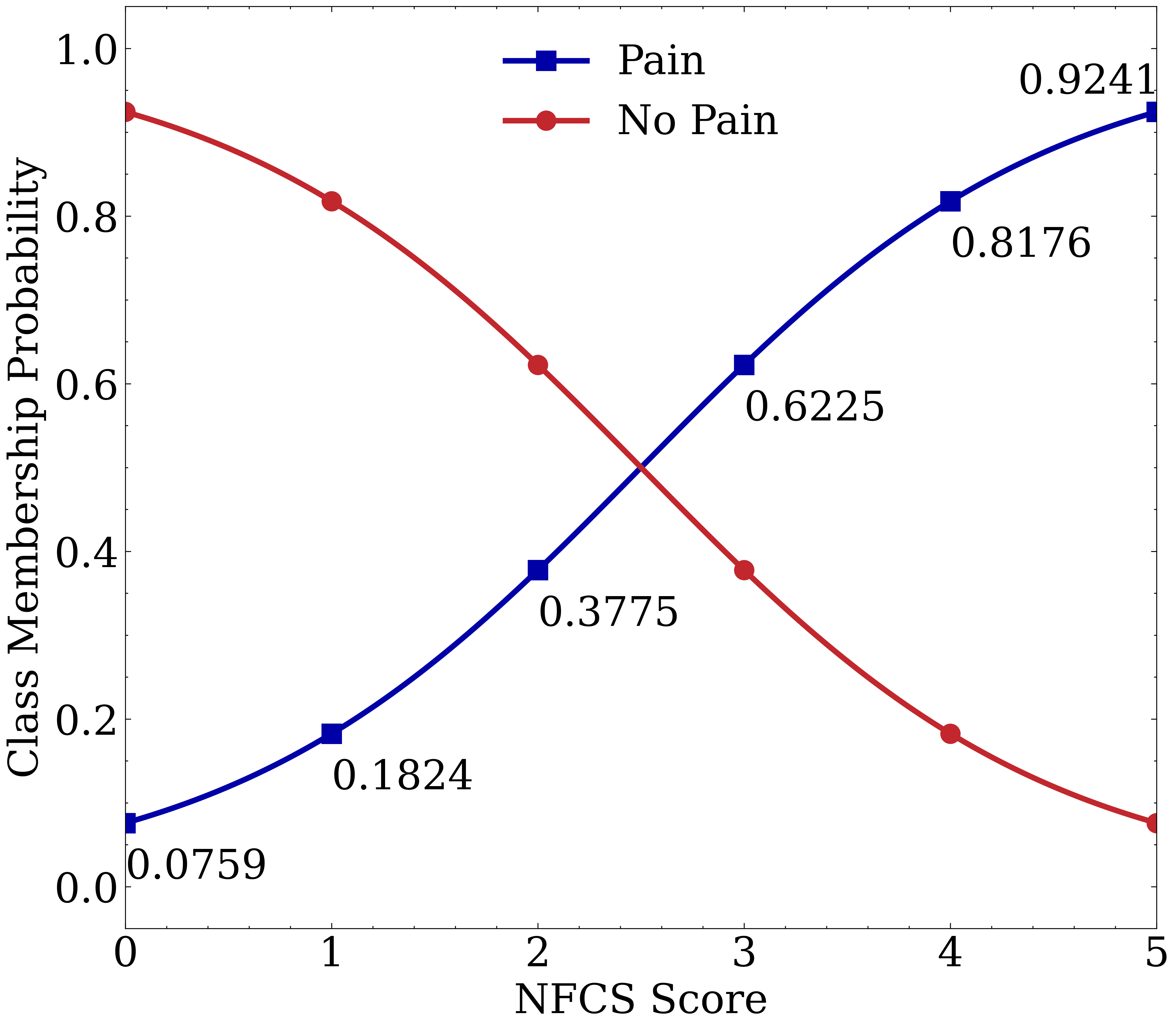}
    \caption{NFCS soft labels.}
    \label{soft_label}
  \end{subfigure}
  \hfill
  \begin{subfigure}[b]{0.47\textwidth}
    \centering
    \includegraphics[width=\textwidth]{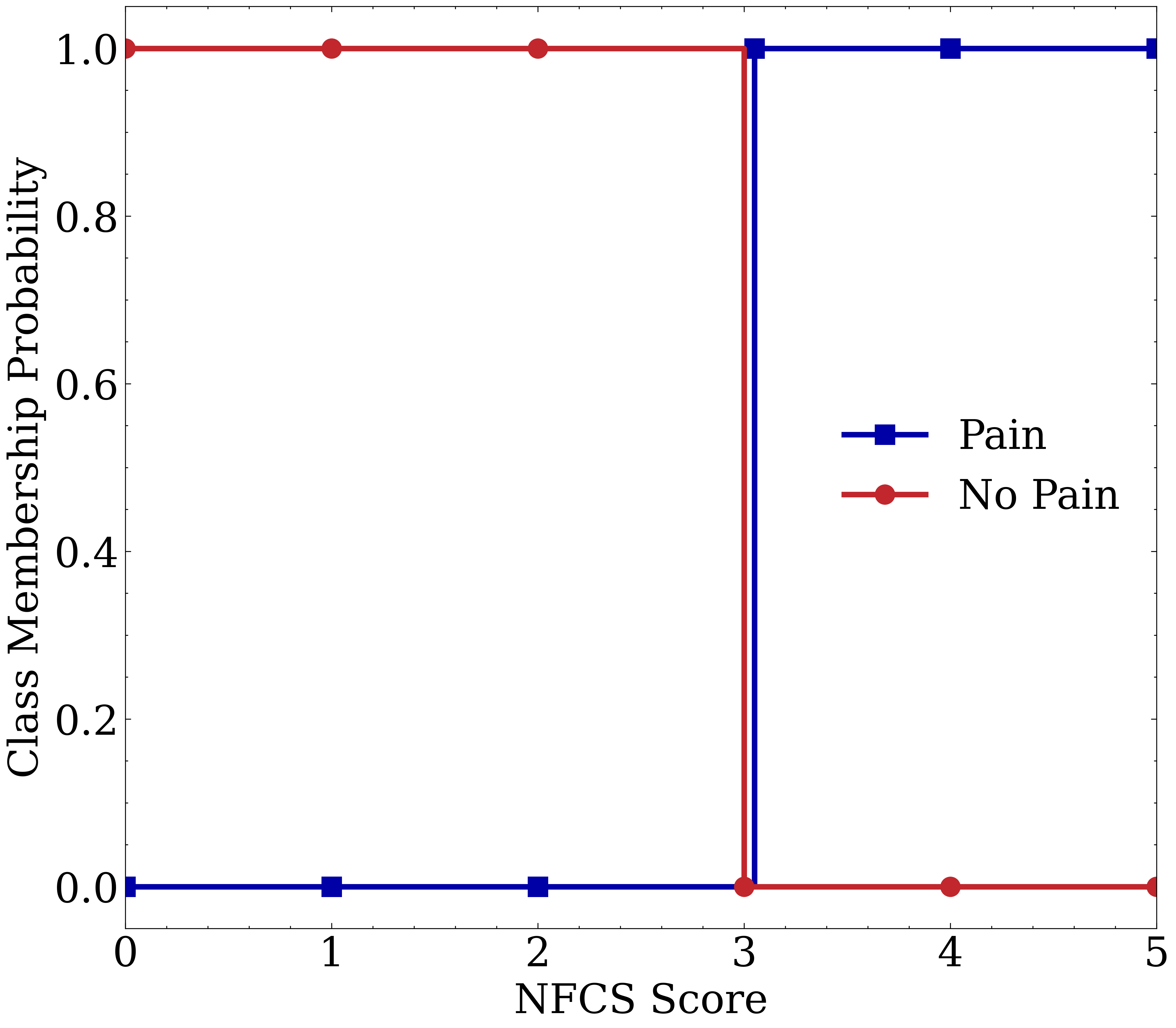}
    \caption{NFCS hard labels.}
    \label{hard_label}
  \end{subfigure}
  \caption{NFCS mapping to soft and hard labels.}
  \label{class_membership}
\end{figure}

The NFCS aims to quantify the presence of 8 facial action units (FAUs) related to pain facial expressions \cite{grunau1987pain}. However, the UNIFESP dataset considered only 5 FAUs due to acquisition limitations \cite{heiderich2015neonatal}. A $NFCS \ge 3$ indicates a facial expression of pain \cite{grunau1987pain}, which is easily mapped to a hard label as seen in Figure \ref{hard_label}. For mapping to a soft label, the sigmoid in Equation \ref{sigmoid_NFCS} was used, as it offers a degree of smoothness in the probability distribution and provides normalized probabilities between 0 and 1. To assure the sigmoid stays in the range (0-5) of the NFCS scores available in the UNIFESP dataset we added the value 2.5:

\begin{equation}
 S(NFCS) = (1+e^{-NFCS+2.5})^{-1}.
\label{sigmoid_NFCS}
\end{equation}

Equation \ref{sigmoid_NFCS} maps the NFCS score to the \textit{“Pain”} class membership probability, while the complement is used for the \textit{“No Pain”} class. The final soft label is made of $[1 - S(NFCS), S(NFCS)]$ (Figure \ref{soft_label}).

Using the NFCS as a soft label, we encourage the output neuron of the \textit{“Pain”} class to be more confident in images with more FAUs of pain. In contrast, the output neuron of the \textit{“No Pain”} class will be more confident when there are few or no FAUs of pain. It is important to acknowledge that the availability of NFCS scores was limited to the UNIFESP dataset. Therefore, to ensure consistency in our evaluations and fair comparisons with other hyperparameters, we excluded the iCOPE images from the original 10 training folds. However, the iCOPE images remained a part of the test folds.

\section{Results}
As we evaluated each hyperparameter individually, some were kept fixed when not being assessed, based on \cite{zamzmi2019pain}: learning rate ($1e^{-4}$), total number of epochs (100), and batch size (16). Model checkpoints were created if the test loss value decreased, and we only selected the lowest loss value checkpoint for evaluation. 
\subsection{Metrics} \label{metrics}
Changing image size and optimizer did not improve F1 Score. The LSR improved F1 Score in $+4.21\%$ when using $\epsilon = 0.3$. Cosine Annealing presented the best results to the learning rate schedulers, with an improvement of $+0.84\%$. Furthermore, increasing the number of epochs led to a notable improvement of $+2.99\%$. Surprisingly, even with $30\%$ fewer images for training, the proposed NFCS soft label achieved the same F1 Score as the Original N-CNN $74.10\%$, demonstrating the positive impact of integrating probabilities
during training.

Table \ref{metrics_table} presents the classification metrics of the test folds for the Original and Tuned N-CNN, with standard deviations. All metrics are statistically significant by the \textit{t-test} ($p < 0.05$), highlighting that hyperparameter optimization alone can lead to improvements in classification results without altering the N-CNN architecture.

\begin{table}[!t]
\centering
\caption{Hyperparameters search space and selected values.}\label{hyper_table}
\begin{tabular}{l|l|l|c|c}
\hline
\textbf{Hyperparameter} & \multicolumn{1}{c|}{\textbf{Search Space}}         & \multicolumn{1}{c|}{\textbf{Selected}} & \textbf{F1 Score} & \textbf{$\Delta$} \\ \hline
Image Size              & $64 \times 64$, $120 \times 120$, $224 \times 224$ & $120 \times 120$        & 74.10\%           & $0\%$               \\
Optimizer               & Adam, Adagrad, RMSProp, SGD                        & RMSprop                 & 74.10\%           & $0\%$               \\
Epochs                  & 50 to 120                                          & 120                     & 77.09\%           & $$+2.99\%$$           \\
Label Smoothing         & 0.1, 0.3, 0.5 and NFCS Soft Label                  & 0.3                     & 78.31\%           & $$+4.21\%$$          \\
Schedulers              & Step, Exponential,                 & Cosine            & 74.94\%           & $$+0.84\%$$         \\
                        & Cosine Annealing                       & Annealing  &  & \\
\hline
\end{tabular}
\end{table}

\begin{table}[!t]
\centering
\caption{Classification metrics for Original and Tuned N-CNN.}\label{metrics_table}
\begin{tabular}{c|c|c|c|c}
\hline
\textbf{Model}    & \textbf{Accuracy}         & \textbf{F1 Score}       & \textbf{Precision}        & \textbf{Recall} \\
\hline
Original & 78.69\% ± 5\% & 74.10\% ± 6\% & 83.17\% ± 8\%  & 68.96\% ± 9\% \\
Tuned       & 82.97\% ± 5\% & 79.74\% ± 6\% & 85.63\% ± 8\% & 75.43\% ± 9\% \\
\hline
$\Delta$  & +4.28\%          & +5.64\%          & +2.46\%          & +6.47\%         \\  
\hline
\end{tabular}
\end{table}
\subsection{Reliability}
\begin{figure}[!t]
  \centering
  \begin{subfigure}[b]{0.47\textwidth}
    \centering
    \includegraphics[width=\textwidth]{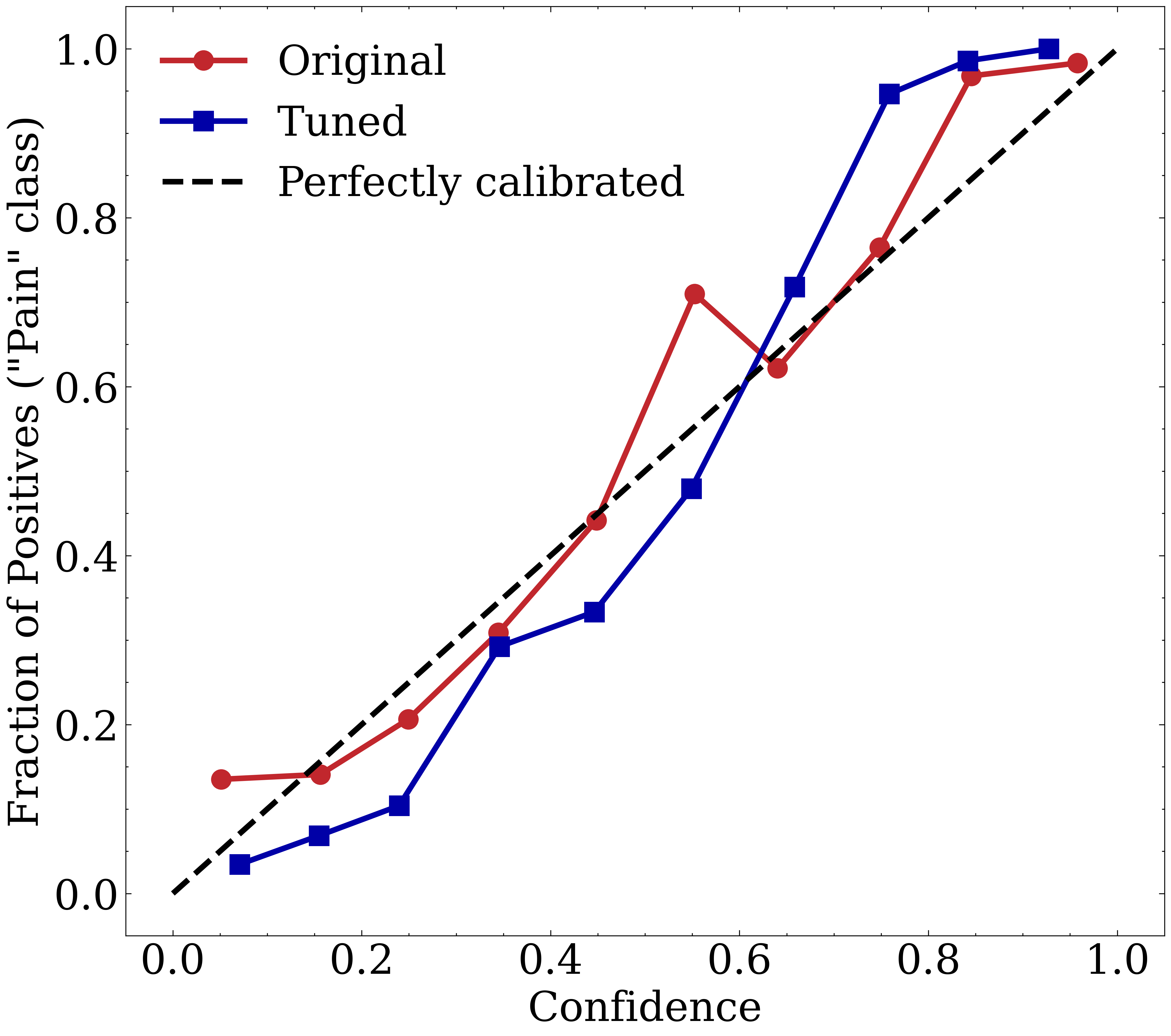}
    \caption{Calibration curves.}
    \label{calibration_curve}
  \end{subfigure}
  \hfill
  \begin{subfigure}[b]{0.47\textwidth}
    \centering
    \includegraphics[width=\textwidth]{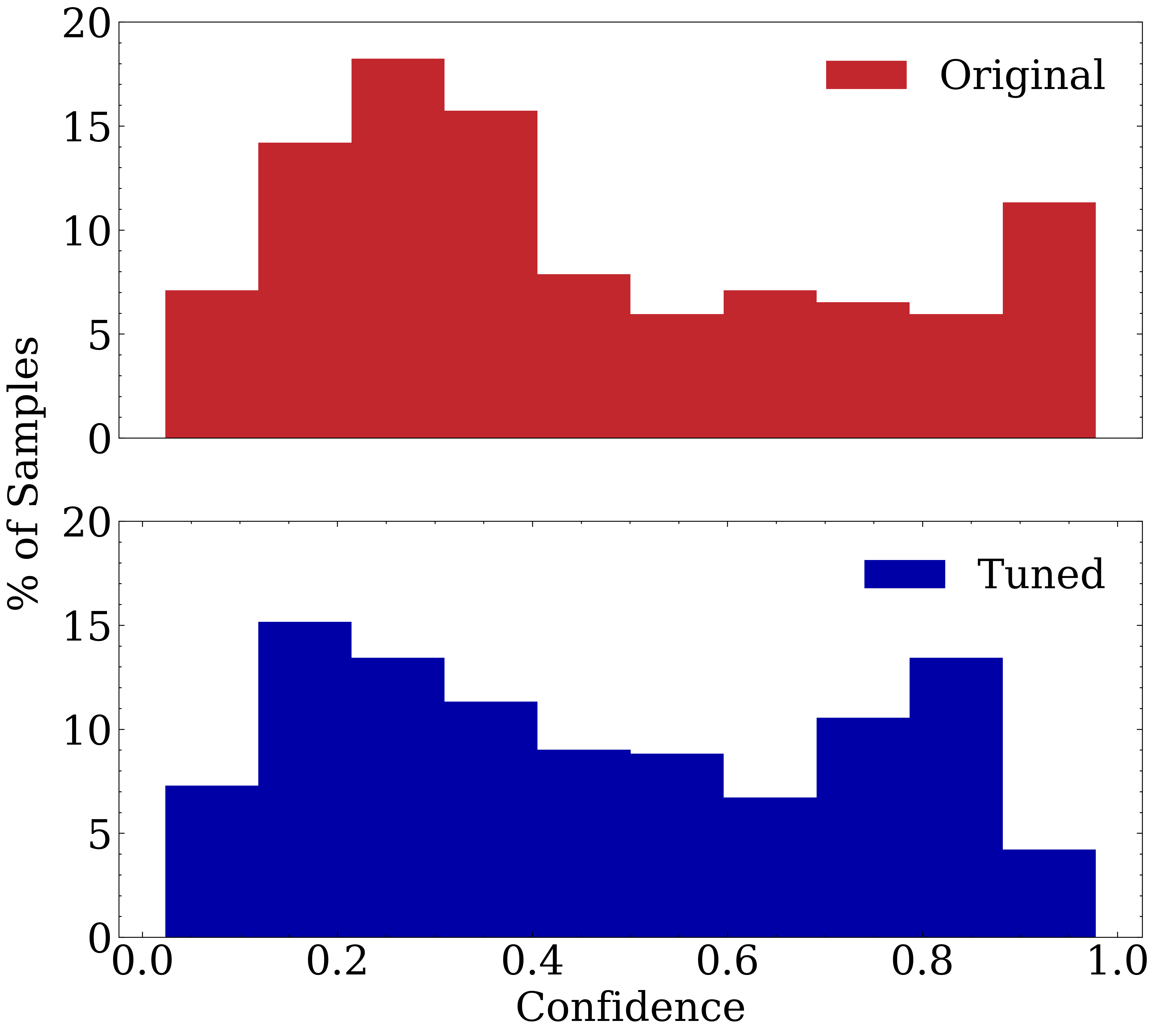}
    \caption{Confidences histogram.}
    \label{conf_hist}
  \end{subfigure}
  \caption{Calibration results of the Original and Tuned N-CNN.}
  \label{calibration_fig}
\end{figure}

Despite achieving higher classification metrics, the Tuned N-CNN did not exhibit the same level of calibration as the Original N-CNN. Specifically, the calibration curve of the Tuned N-CNN displayed an “S” shape, indicating overestimation of predictions with low confidences and underestimation of predictions with high confidences (Figure \ref{calibration_curve}). This difference in calibration is confirmed by the ECE metric \cite{naeini2015obtaining}, with the Tuned N-CNN yielding an $ECE = 0.091$, approximately three times greater than the Original N-CNN's $ECE = 0.035$.

As shown in Figure \ref{calibration_curve}, of the predicted samples for which the model exhibits confidence around $20\%$, less than $10\%$ of them were actually in the \textit{“Pain”} class. On the other hand, almost all samples with $80\%$ confidence were of the \textit{“Pain”} class, indicating that the Tuned N-CNN has a lower confidence in this interval. Conversely, when considering the Original N-CNN, the model exhibits limited confidence in the $50\%$ to $60\%$ range, indicating challenges in confidently classifying samples within the threshold region of the \textit{“Pain”} class.

The low confidence phenomenon in Tuned N-CNN, is further depicted in Figure \ref{conf_hist}, showing how the percentage of samples with confidences over 90\%, from Original N-CNN, shifted to a less confident range due to the LSR technique which tends to reduce model overconfidence.

\subsection{Explainability}
Figures \ref{GC_XAI} and \ref{IG_XAI} depict images correctly classified by both models, providing insights into the explainability improvements when accessing the Tuned N-CNN. For example, the Tuned N-CNN focused on regions of the NBs face, ignoring artifacts such as clothing (Figure \ref{GC_XAI}d). In addition, it was found that the $IG_{Tuned}$ acquired a more uniform distribution across the NBs face, unlike the $IG_{Original}$, which gave more attention to the eyes and mouth. As for the \textit{“No Pain”} images, $GC_{Tuned}$ concentrated on the forehead and cheeks (Figure \ref{GC_XAI}e), contrasting with $GC_{Original}$, which did not obtain a central area of focus.

Our results revealed a correlation between XAI and the predicted confidence in both Original and Tuned N-CNNs. Confidences ranging from $40\%$ to $60\%$ were observed to explain features not related to the NB face, as it can be seen in Figure \ref{GC_XAI}c and \ref{GC_XAI}f, that is, an overconfidence in the \textit{“Pain”} class. This interpretation was also observed for images that both models incorrectly classified. In contrast, confidences above $80\%$ or below $20\%$ in the \textit{“Pain”} class that were correctly classified presented an attribution mask deemed more relevant to clinical practice. Moreover, both $IG$ and $GC$ results mainly gave more importance to the mouth region, which agrees with one of the observed regions by neonatologists \cite{carlini2020visual}.

\def \picsize {0.28\textwidth}
\begin{figure}[!b]
\centering
\begin{tabular}{c||c c c}
\hline
\textbf{Model} & \textbf{Pain} & \textbf{No Pain} & \textbf{No Pain} \\ \hline
\multirow{9}{*}{Original} & 
\multirow{2}{*}{\includegraphics[height=\picsize,width=\picsize]{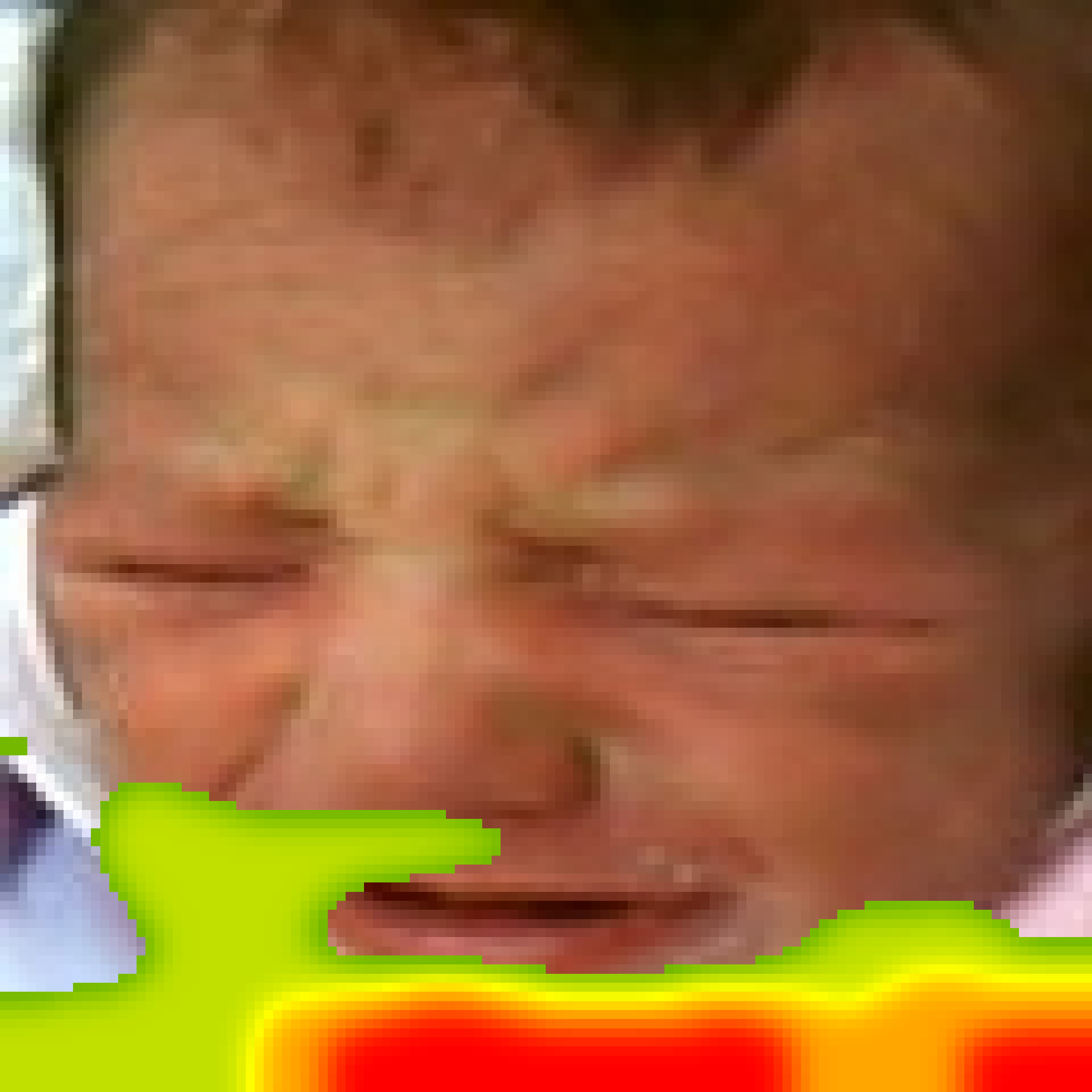}} & 
\multirow{2}{*}{\includegraphics[height=\picsize,width=\picsize]{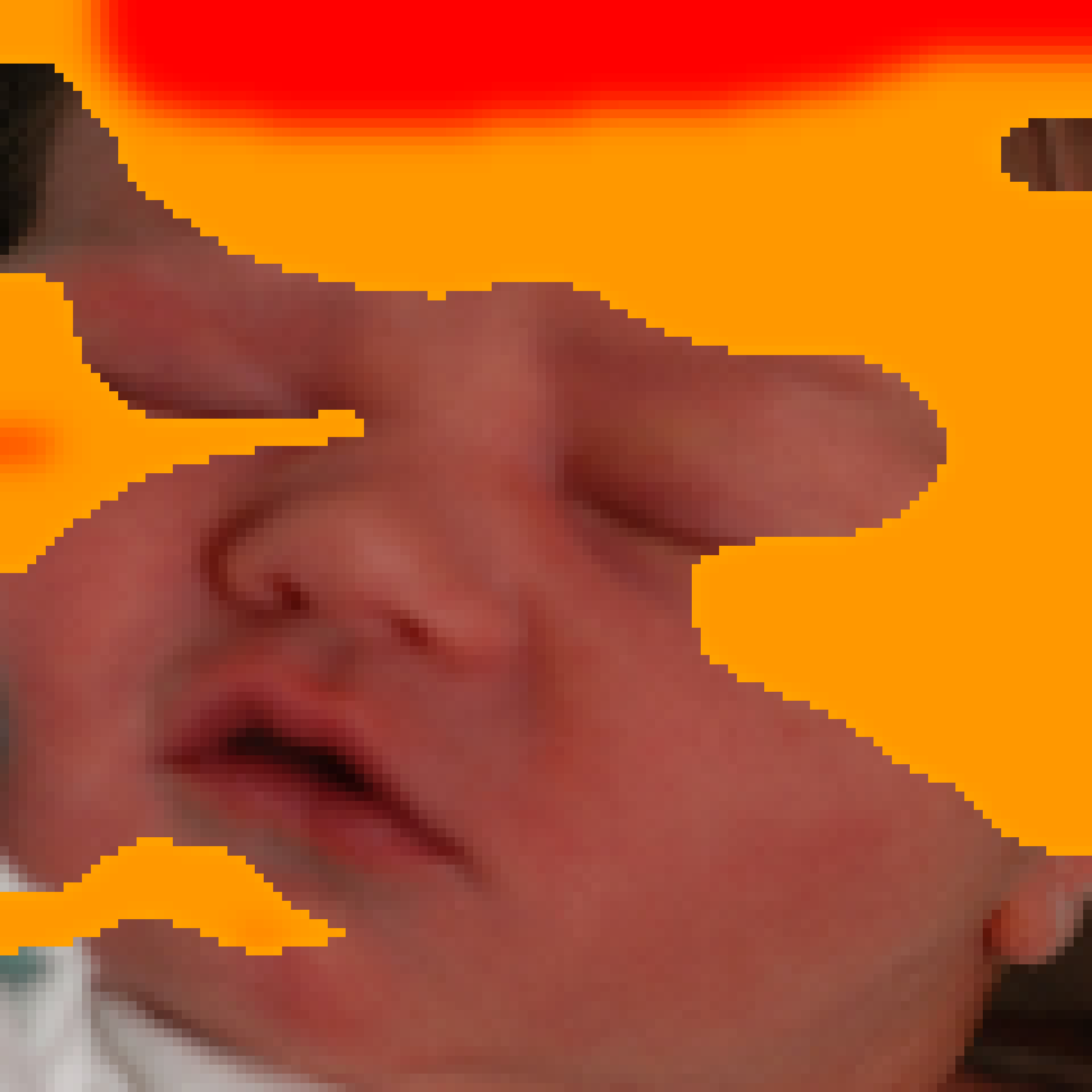}} & 
\multirow{2}{*}{\includegraphics[height=\picsize,width=\picsize]{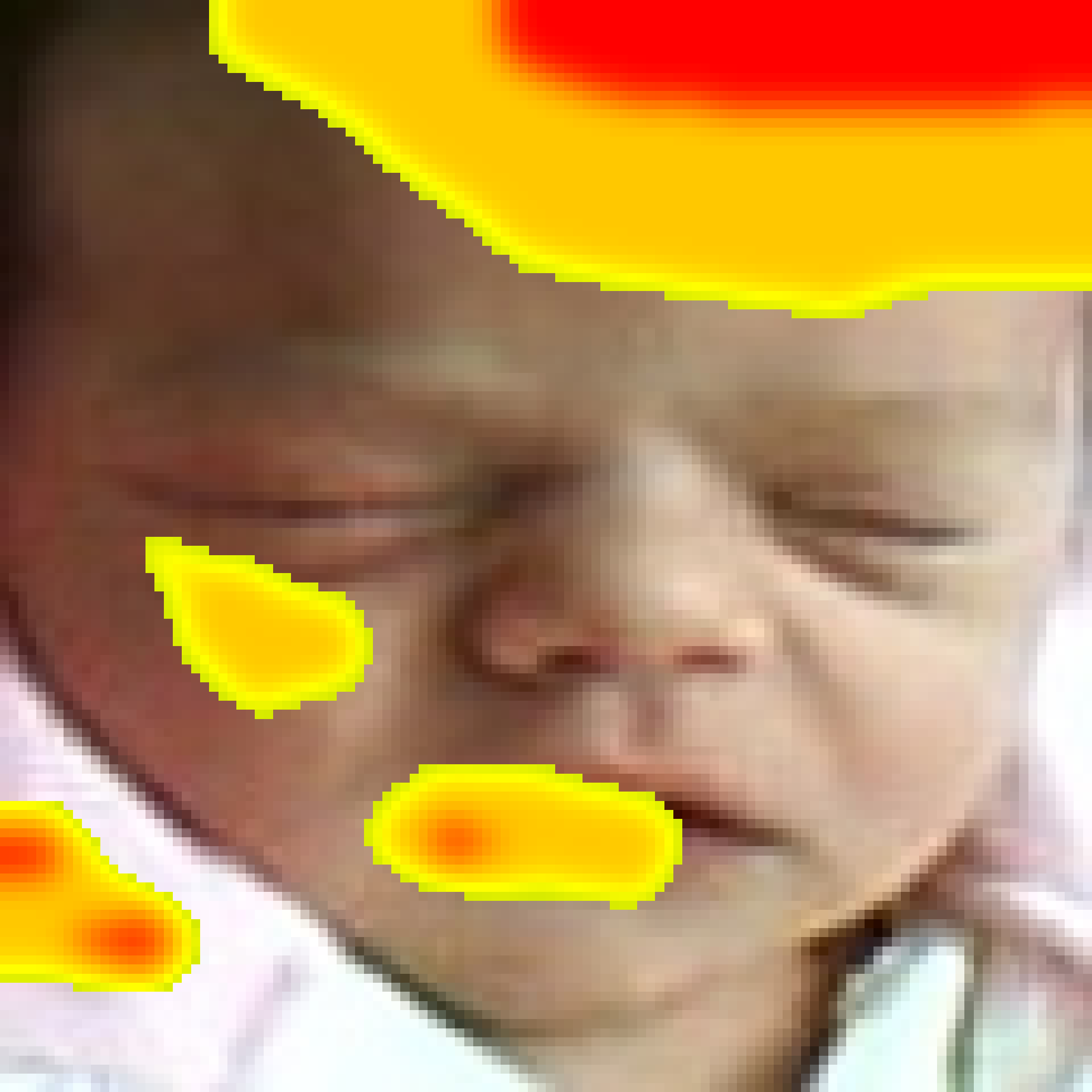}}\\
 \cr \\
 \cr \\
 \cr \\
 \cr \\
& (a) Conf. 50.50\% & (b) Conf. 43.96\% & (c) Conf. 43.29\% \\
 
\multirow{9}{*}{Tuned} &
\multirow{2}{*}{\includegraphics[height=\picsize,width=\picsize]{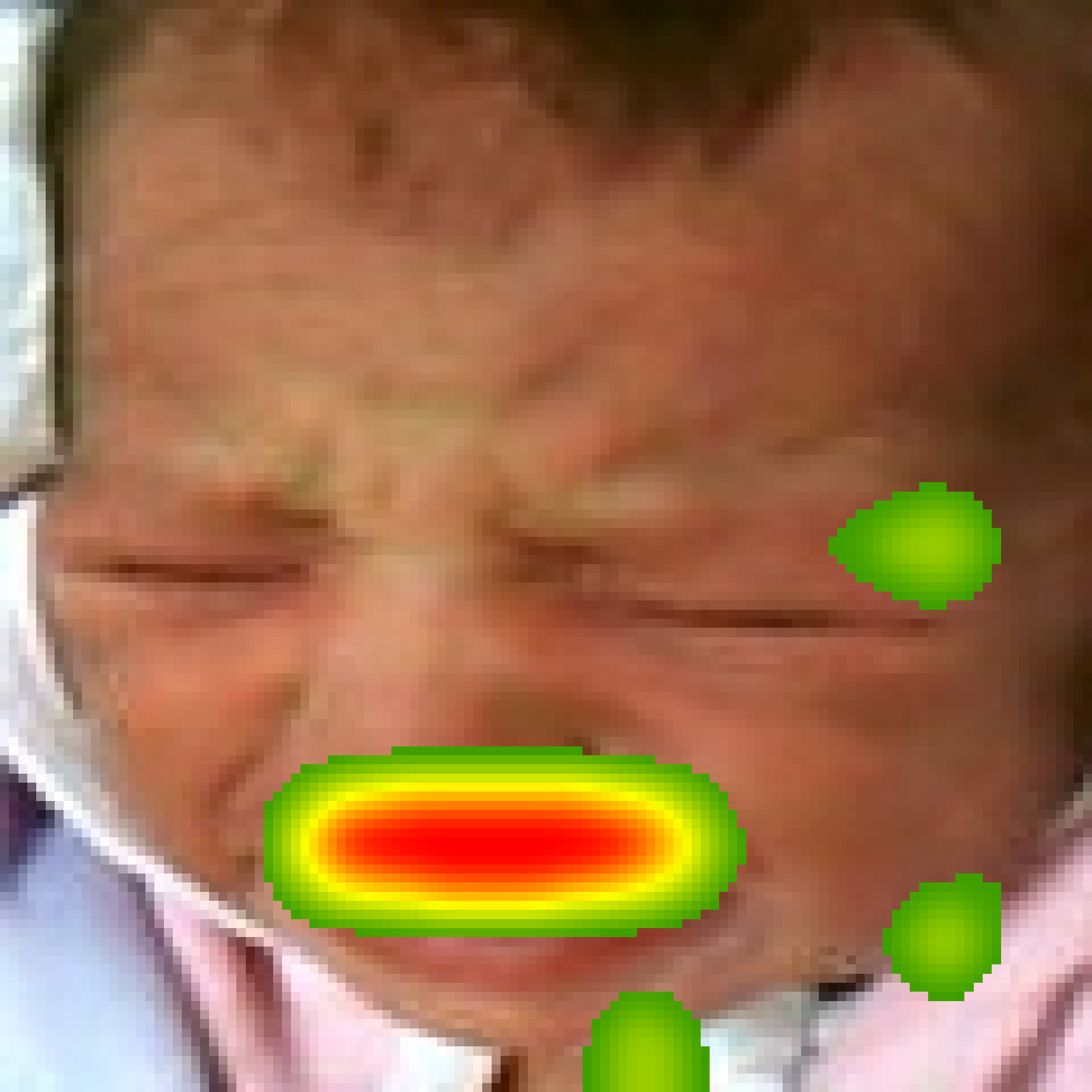}} &
\multirow{2}{*}{\includegraphics[height=\picsize,width=\picsize]{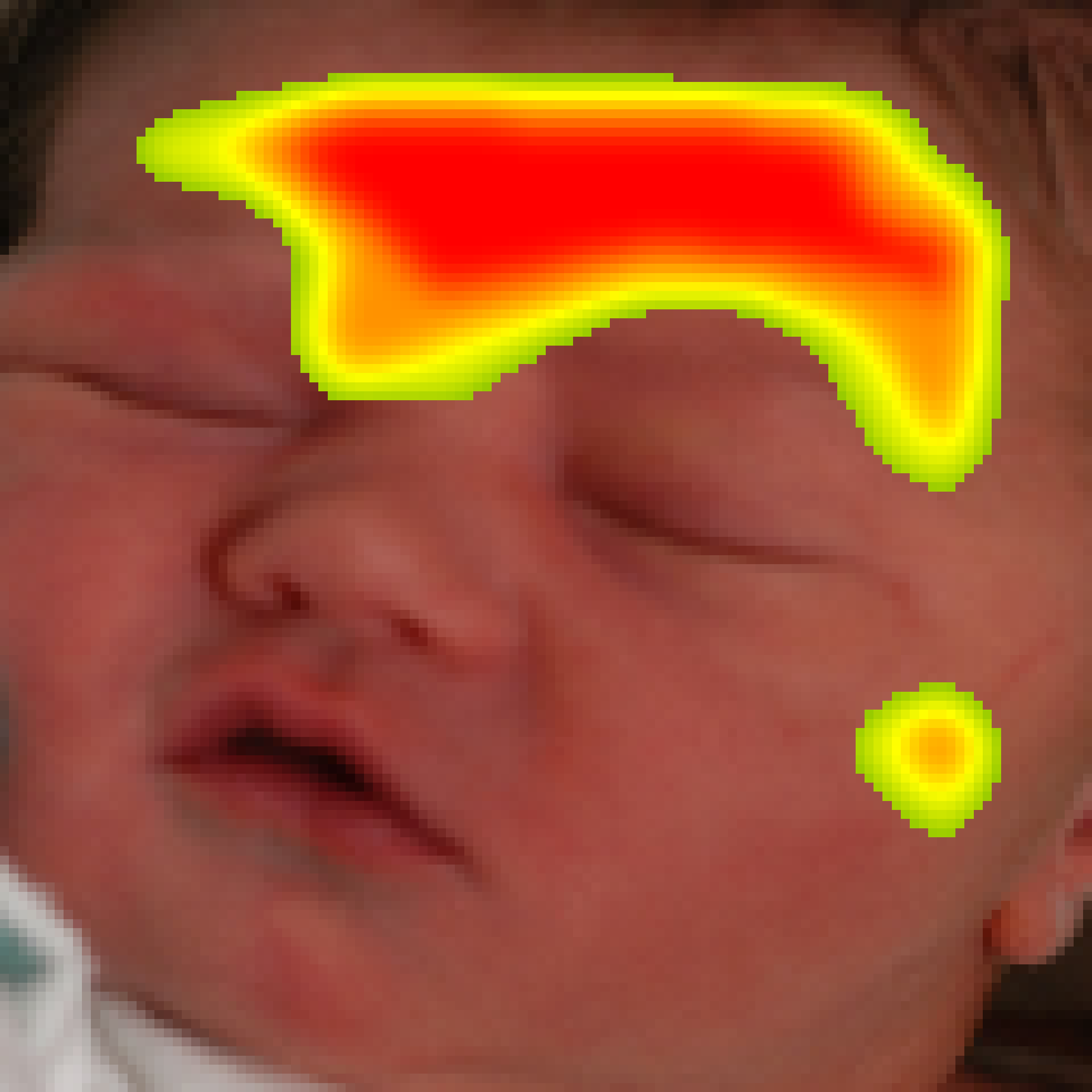}} &
\multirow{2}{*}{\includegraphics[height=\picsize,width=\picsize]{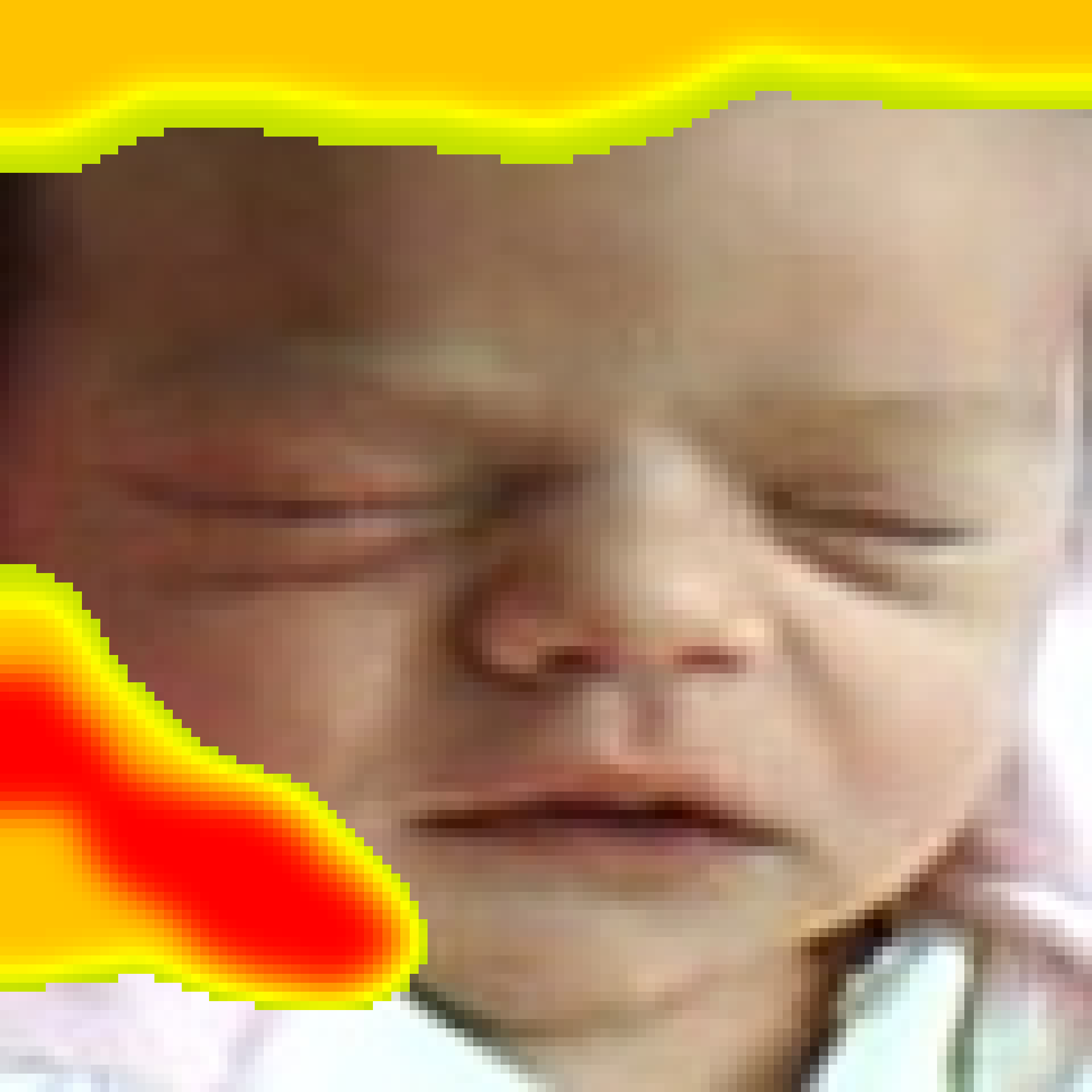}}\\
 \cr \\
 \cr \\
 \cr \\
 \cr \\
& (d) Conf. 89.58\% & (e) Conf. 13.32\% & (f) Conf. 39.96\%\\

\hline
\end{tabular}
\caption{$GC$ attribution masks and predictions confidence in the \textit{“Pain”} class.}
\label{GC_XAI}
\end{figure}

\begin{figure}[!t]
\centering
\begin{tabular}{c||c c c}
\hline
\textbf{Model} & \textbf{Pain} & \textbf{No Pain} & \textbf{No Pain} \\ \hline
\multirow{9}{*}{Original} & 
\multirow{2}{*}{\includegraphics[height=\picsize,width=\picsize]{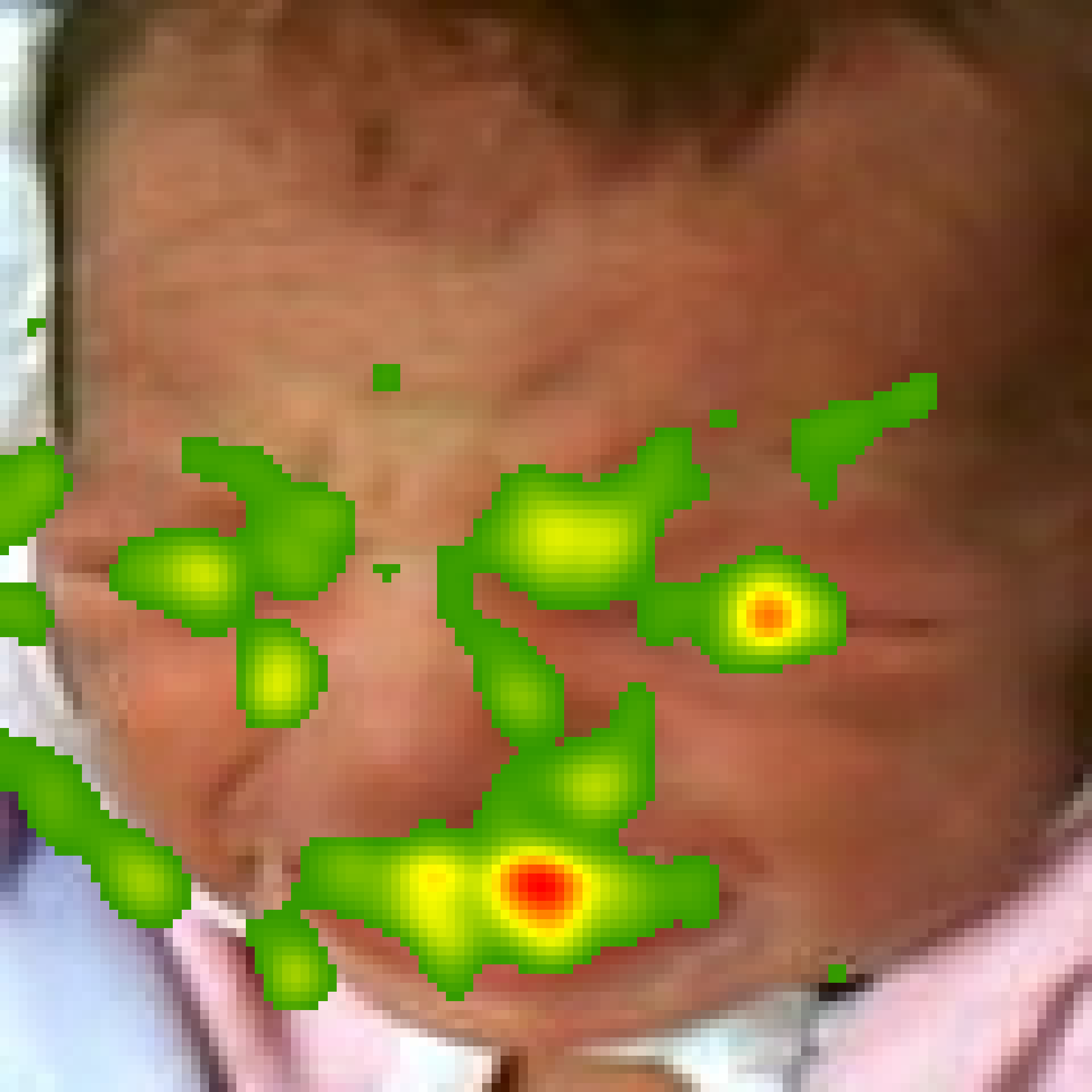}} &
\multirow{2}{*}{\includegraphics[height=\picsize,width=\picsize]{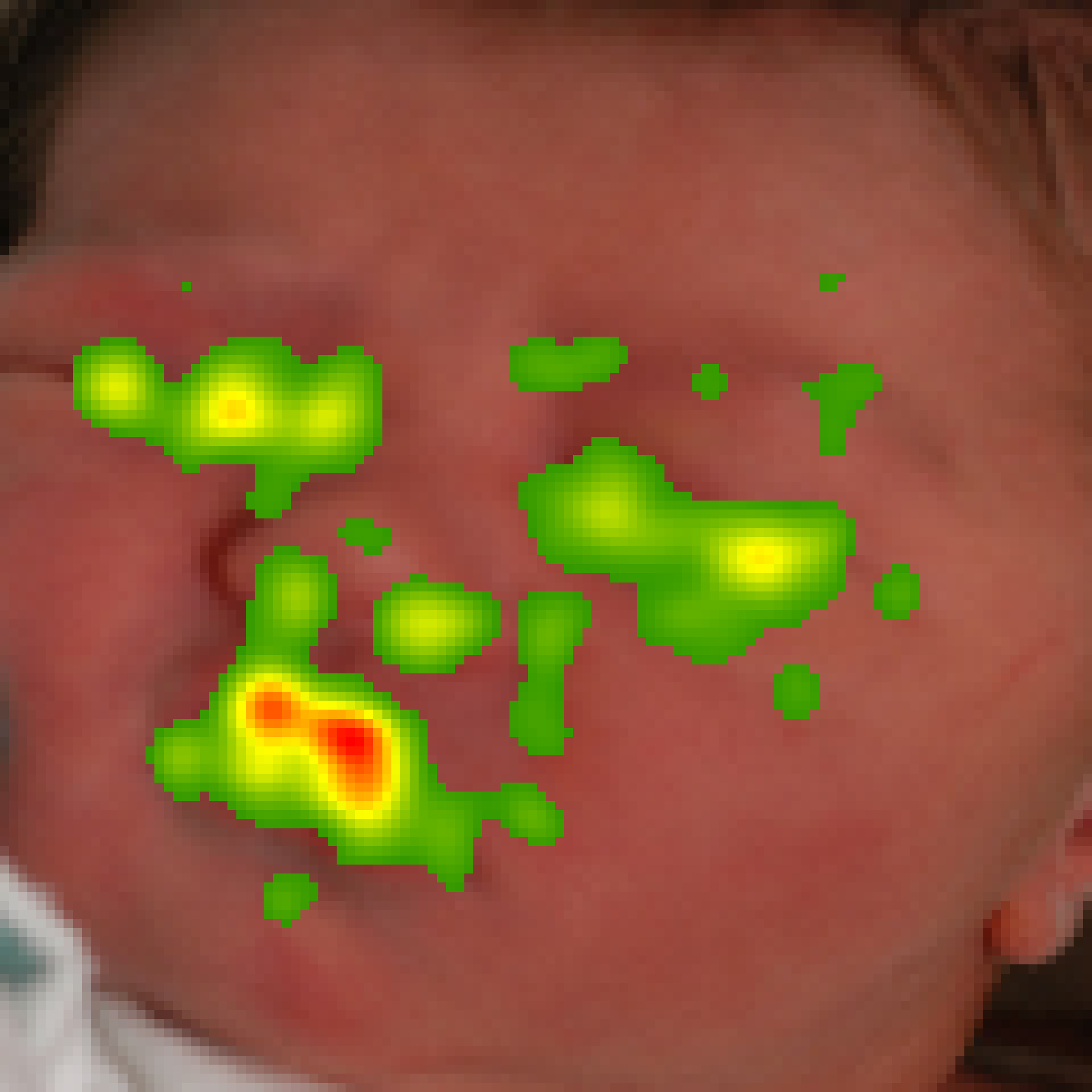}} &
\multirow{2}{*}{\includegraphics[height=\picsize,width=\picsize]{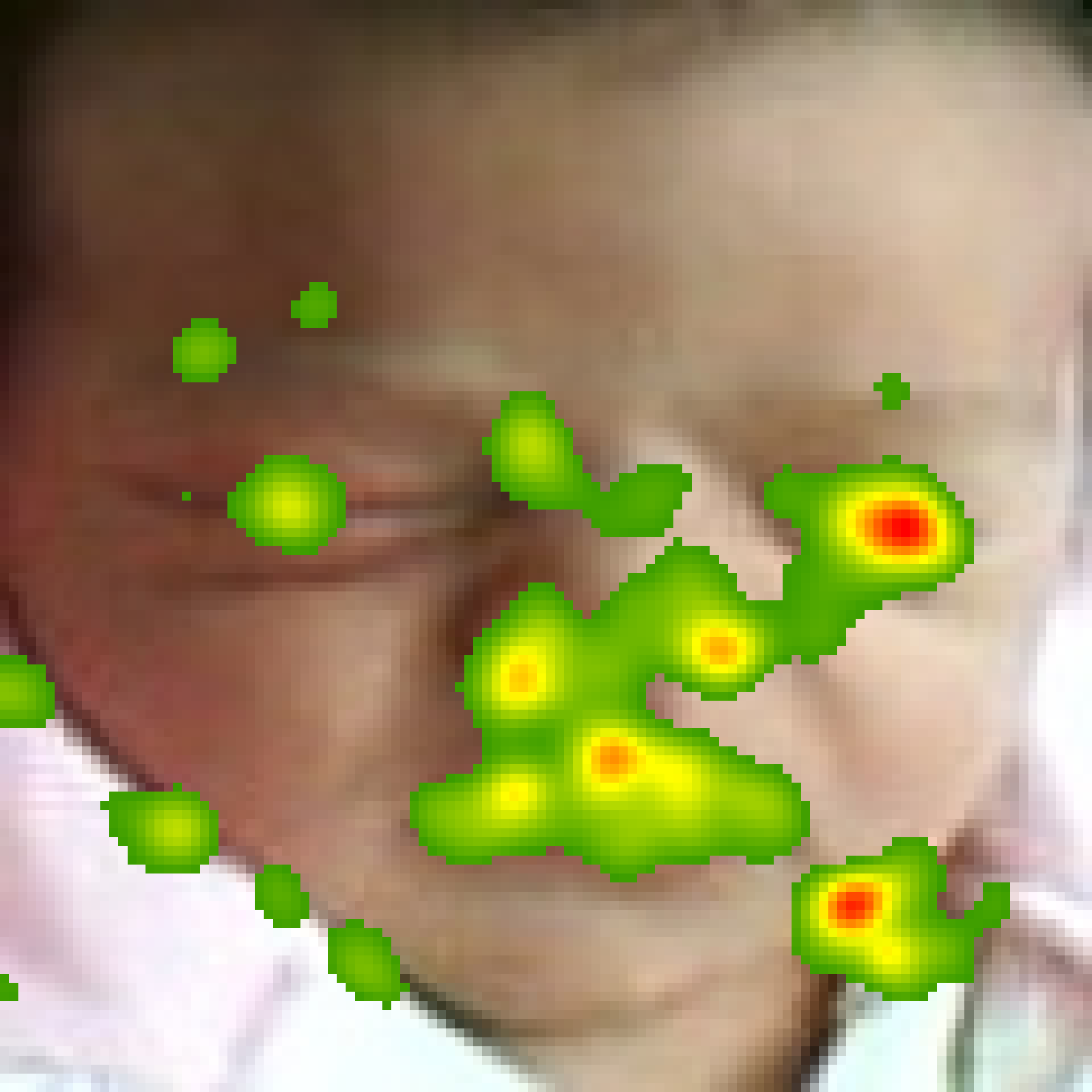}} \\
 \cr \\
 \cr \\
 \cr \\
 \cr \\
& (a) Conf. 50.50\% & (b) Conf. 43.96\% & (c) Conf. 43.29\% \\
 
\multirow{9}{*}{Tuned} &
\multirow{2}{*}{\includegraphics[height=\picsize,width=\picsize]{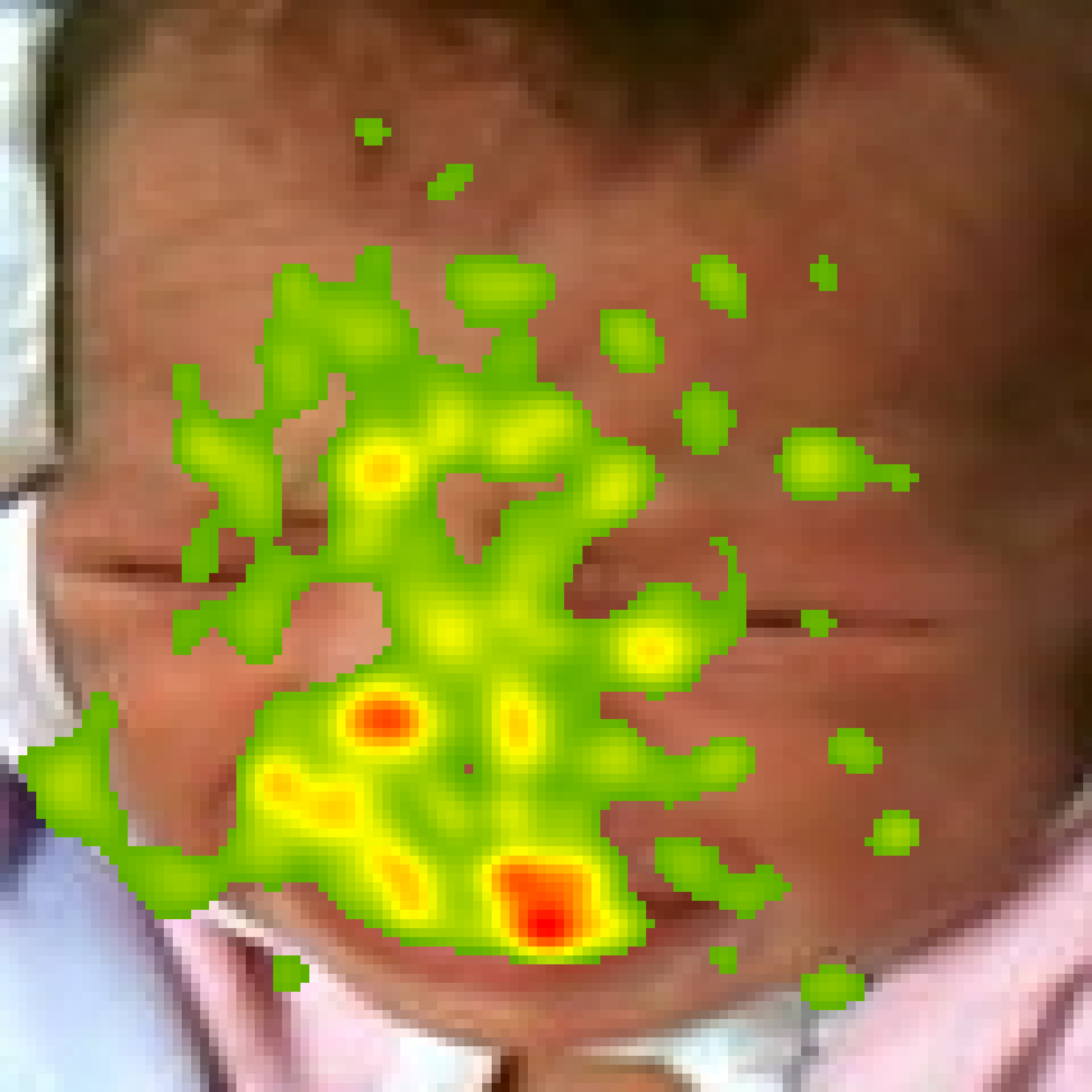}} &
\multirow{2}{*}{\includegraphics[height=\picsize,width=\picsize]{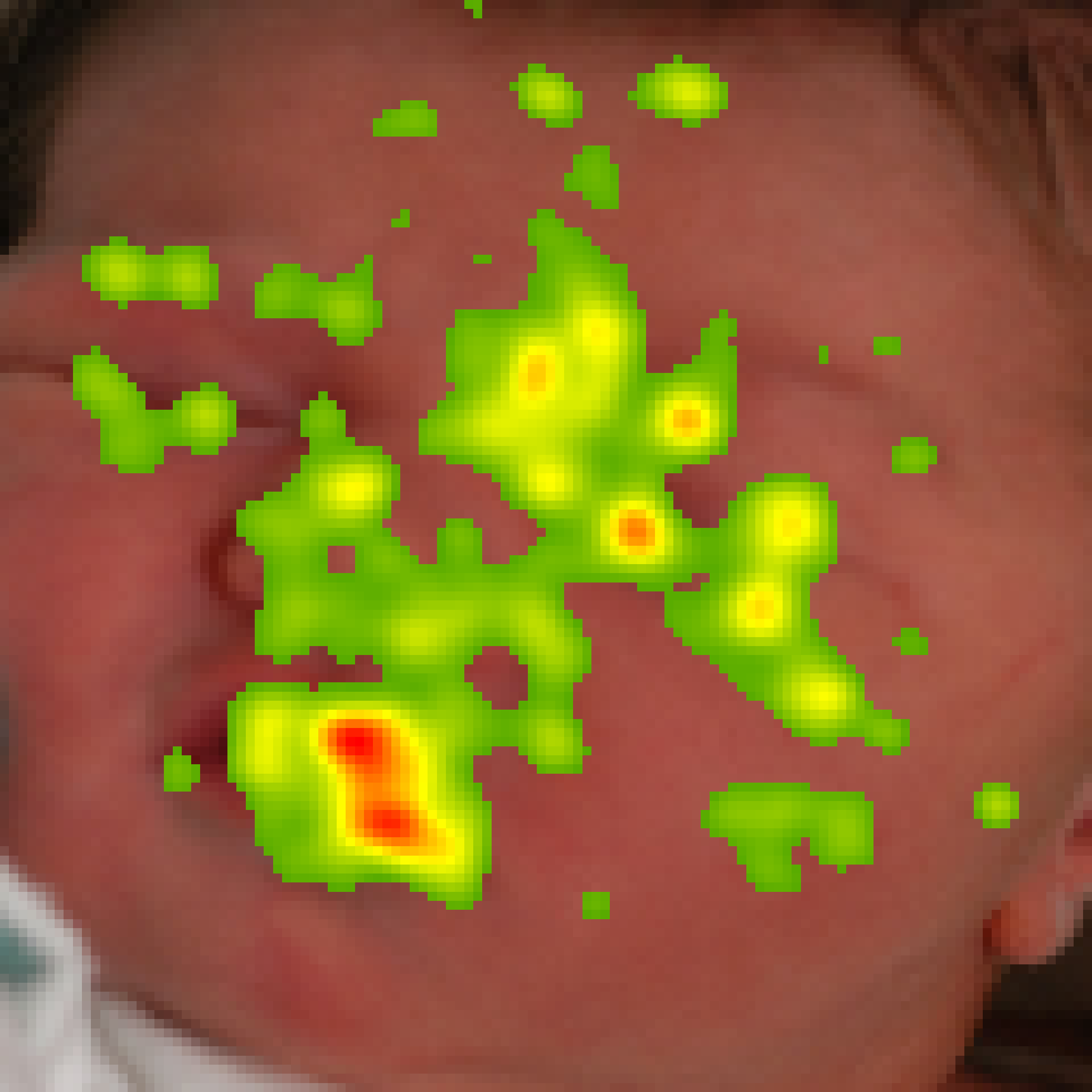}} &
\multirow{2}{*}{\includegraphics[height=\picsize,width=\picsize]{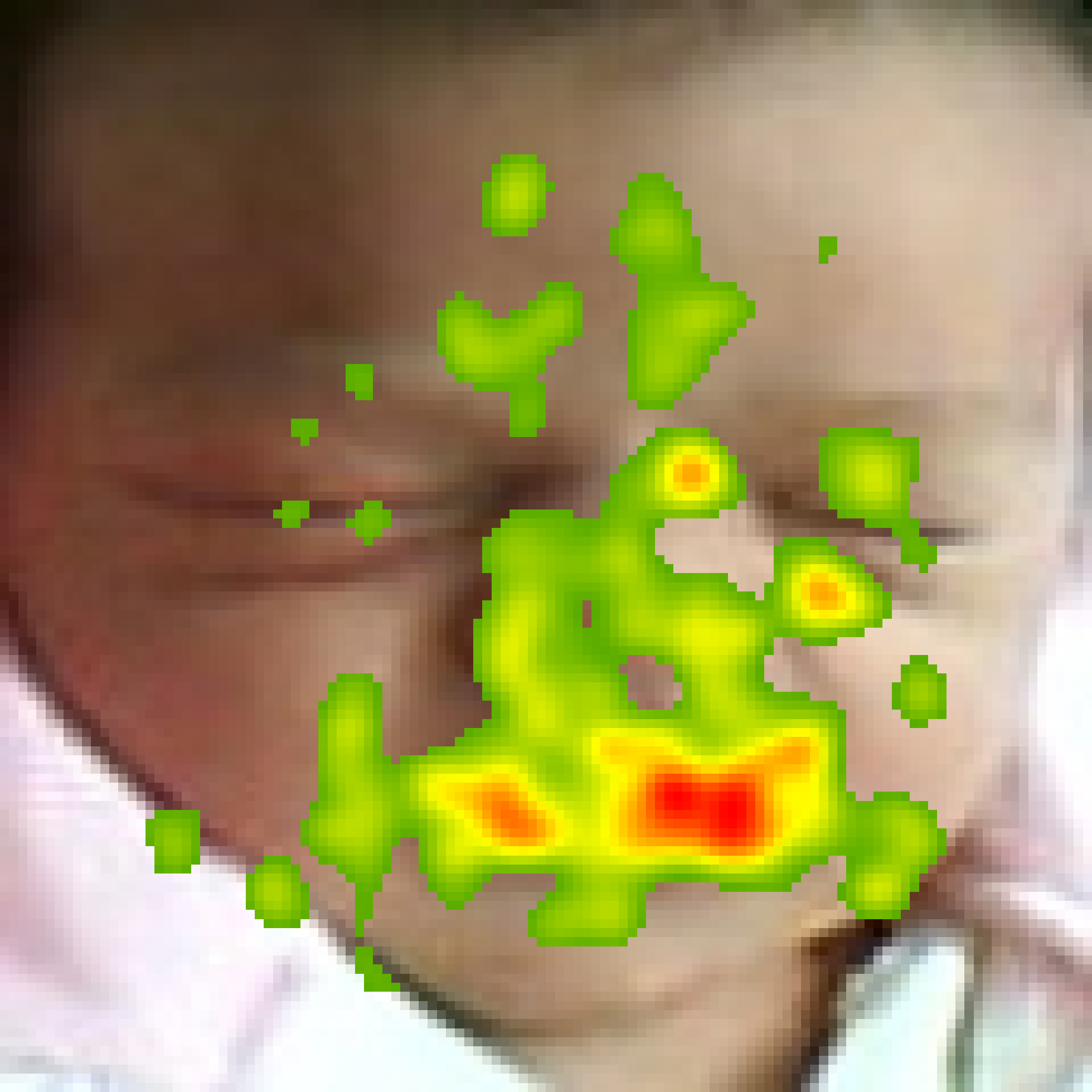}} \\
 \cr \\
 \cr \\
 \cr \\
 \cr \\
& (d) Conf. 89.58\% & (e) Conf. 13.32\% & (f) Conf. 39.96\%\\

\hline
\end{tabular}
\caption{$IG$ attribution masks and predictions confidence in the \textit{“Pain”} class.}
\label{IG_XAI}
\end{figure}

\section{Discussion}
Based on our observations, training with soft labels instead of hard labels proved beneficial to the N-CNN classification metrics, which has not been extensively explored in the existing literature. We believe that the NFCS soft label proposed here can surpass the best LSR result found if iCOPE NFCS scores were available, emphasizing the necessity for standardization across neonatal pain scales in datasets, promoting consistency and comparability in future research. This technique can also be applied to various clinical pain scales that quantify the presence or absence of pain, not only to NFCS.

Regarding reliability, calibrated predictions are essential to accurately estimate the associated risks in critical settings like in NICU. For instance, predictions with $90\%$ confidence in the \textit{“Pain”} class should prompt immediate action as the NB has a real $90\%$ chance of being in pain, while predictions with $60\%$ confidence require more careful analysis by healthcare professionals, as the AI model exhibits uncertainty in its prediction for the specific situation of the NB. The fact that the calibration of the Tuned N-CNN has worsened raises an important question: to what extent does improving classification metrics benefit clinical practice? While a $+5\%$ increase in the Tuned N-CNN metrics may lead to more accurate automatic predictions, healthcare professionals may only rely on these predictions if the model's decision-making process is explainable and aligned with the actual probability of event occurrence.

The use of $IG$ on the Tuned N-CNN revealed that more pixels contributed to the final prediction. We believe that this increase in relevant pixels were a contributing factor to the Tuned N-CNN achieving higher metrics. However, the N-CNN still has some explainability limitations when compared to other Deep-Learning models \cite{coutrin2022}, due to the shallow architecture which may limit its capacity to capture complex patterns and nuances in neonatal pain assessment. Nevertheless, by visualizing these attribution masks, healthcare professionals can better understand the model's reasoning and gain trust in its predictions.

\section{Conclusion}
In this work, we revisited the N-CNN model by optimizing its hyperparameters and evaluating how tuning affects its classification metrics, explainability and reliability. Although we observed improvements in classification metrics and explainability, these improvements did not simply translate to calibration performance. An uncalibrated model applied in clinical practice can lead to a lack of trust by the health professional in the model's predictions, resulting in the rejection of the technology. While enhancements in traditional metrics are notable, achieving reliability and explainability in any AI model intended for the health sector is imperative to ensure reliable and actionable predictions.

We believe that the new approach proposed here of using the NFCS as a soft label might disclose a new paradigm in training classification models for neonatal pain analysis based on facial expressions. As future work, we intend to explore soft labels based on clinical pain scales and calibration methods, ensuring that confidences align with the actual frequency of both pain and no-pain events. Overall, we understand that it is crucial to address the issue of uncalibrated confidences as risk estimators in the medical field, not only to prevent erroneous automatic decision-making and foster trust among healthcare professionals, but also to facilitate the responsible and effective use of AI in clinical practice.

\subsubsection{Acknowledgments}
This study was financed in part by CAPES (Finance Code 001) and FAPESP (2018/13076-9). 

\bibliographystyle{splncs04}
\bibliography{referencias}

\end{document}